%% file: acl_latex.tex
\documentclass[11pt]{article}

\usepackage[final]{acl}
\usepackage{times}
\usepackage{latexsym}
\usepackage{latexsym}
\usepackage{amsmath}
\usepackage{algorithm}
\usepackage{algpseudocode}
\usepackage{float}
\usepackage[T1]{fontenc}

\usepackage[utf8]{inputenc}
\usepackage[most]{tcolorbox}
\usepackage{amssymb}   
\usepackage{xcolor}    
\usepackage{booktabs}  
\usepackage{multirow}  
\usepackage{microtype}

\usepackage{inconsolata}

\usepackage{graphicx}

\title{KG-ViP: Bridging \underline{K}nowledge \underline{G}rounding and \underline{Vi}sual \underline{P}erception  in Multi-modal LLMs for Visual Question Answering}

\author{
    \textbf{Zhiyang Li}$^{1,2,}$\thanks{~~Equal contribution.} \quad
    \textbf{Ao Ke}$^{1,2,}$\footnotemark[1] \quad
    \textbf{Yukun Cao}$^{2,3,}$\thanks{~~Corresponding authors.} \quad
    \textbf{Xike Xie}$^{1,2,}$\footnotemark[2] \\
    $^{1}$University of Science and Technology of China, China \\
    $^{2}$Data Darkness Lab, MIRACLE Center, USTC \\
    $^{3}$School of Computer Science and Technology, Xidian University \\
    \texttt{\{lizhiyang, sa21225249\}@mail.ustc.edu.cn} \\
    \texttt{caoyukun@xidian.edu.cn, xkxie@ustc.edu.cn}
}
\usepackage{verbatim}
\usepackage{booktabs}
\usepackage{tcolorbox} 
\usepackage{hyperref}
\tcbuselibrary{breakable}
\usepackage{enumitem}
\usepackage{geometry}
\usepackage{booktabs}  
\usepackage{multirow}   
\usepackage{colortbl}   
\usepackage{xcolor}     
\usepackage{graphicx}   
\usepackage{float}
\usepackage{graphicx}
\usepackage{booktabs}
\usepackage{pifont}
\usepackage{xcolor}

\definecolor{rowblue}{RGB}{236, 246, 255}
\setlist[itemize]{leftmargin=*, nosep}
\setlist[enumerate]{leftmargin=*, nosep}

\usepackage{booktabs}
\usepackage{multirow}
\usepackage{colortbl}
\usepackage{xcolor}
\usepackage{pifont}
\usepackage{enumitem}
\setlist[itemize]{left=0pt}
\definecolor{rowblue}{RGB}{235,245,255}

\usepackage{amsthm}

\begin{document}
\maketitle
\input{Section/Abstract}
\input{Section/Introduction}

\input{Section/Related_Work}

\input{Section/Method}

\input{Section/Experiment}

\input{Section/Conclusions}
\input{Section/Limitations}
\input{Section/Acknowledgements}
\bibliography{KG-ViP}
\input{Section/Appendices}

\end{document}

%% file: Section/Abstract.tex
\begin{abstract}

Multi-modal Large Language Models (MLLMs) for Visual Question Answering (VQA) often suffer from dual limitations: knowledge hallucination and insufficient fine-grained visual perception. Crucially, we identify that commonsense graphs and scene graphs provide precisely complementary solutions to these respective deficiencies by providing rich external knowledge and capturing fine-grained visual details. However, prior works typically treat them in isolation, overlooking their synergistic potential. To bridge this gap, we propose \textbf{KG-ViP}, a unified framework that empowers MLLMs by fusing scene graphs and commonsense graphs. The core of the KG-ViP framework is a novel retrieval-and-fusion pipeline that utilizes the query as a semantic bridge to progressively integrate both graphs, synthesizing a unified structured context that facilitates reliable multi-modal reasoning. Extensive experiments on FVQA 2.0+ and MVQA benchmarks demonstrate that KG-ViP significantly outperforms existing VQA methods.
\end{abstract}

%% file: Section/Introduction.tex
\section{Introduction}

\begin{figure}[t]
    \centering
    \includegraphics[width=\columnwidth]{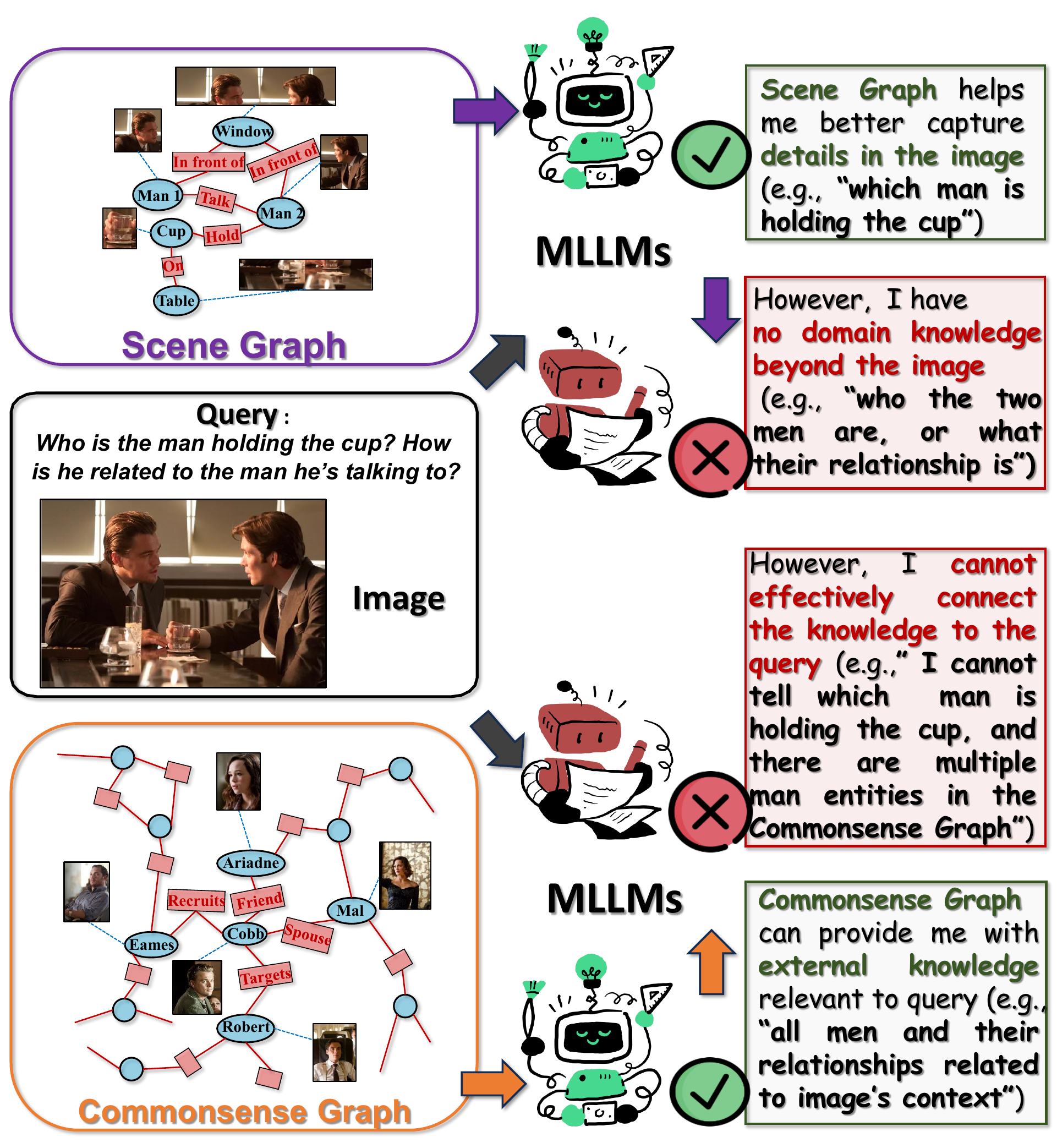} 
    \caption{
      \textbf{Complementary roles of Scene Graphs and Commonsense Graphs.}
    \footnotesize 
     Using a scene from \textit{Inception} as an example: Given the query asking \textit{``Who is the man holding the cup?''}, the \textbf{scene graph} explicitly captures visual relations (e.g., \textit{holding}) yet fails to recognize character identities (\textit{Cobb}, \textit{Robert}). 
    Conversely, the \textbf{commonsense graph} offers semantic facts (e.g., \textit{Cobb targets Robert}) but suffers from grounding ambiguity, as it lacks visual cues to anchor abstract nodes to the specific person in the image. 
    Thus, neither graph alone can support the complete reasoning chain.
    }
    \label{fig:motivation}
    \vspace{-20pt}
\end{figure}

Multi-modal large language models (MLLMs) have demonstrated strong performance on Visual Question Answering (VQA)~\cite{okvqa,MMBench,mlevlm,deliberate}, a task that requires joint reasoning over visual content and textual queries.
However, MLLMs still suffer from dual limitations stemming from deficiencies in both knowledge grounding and visual perception.

From the knowledge perspective, when confronted with knowledge-intensive questions, MLLMs are prone to generating hallucinations~\cite{EOHLVLM,HallusionBench,MMCD}. 
To mitigate this issue, Retrieval-Augmented Generation (RAG)~\cite{fan2024survey} has been widely adopted to incorporate external knowledge. 
In particular, Graph-based RAG~\cite{graphrag,TOG,legoRAG} has recently emerged as a compelling paradigm, where knowledge is organized into structured entities and relations (e.g., \textit{commonsense graphs}), enabling more precise knowledge retrieval and facilitating explicit multi-hop reasoning.

From the perception perspective, prior work~\cite{SGVL,ma2023crepe, MitraCCoT} has revealed that MLLMs exhibit limited visual understanding, particularly in capturing fine-grained relationships and interactions within complex scenes.
To enhance the perceptual capability of MLLMs, a promising direction is to introduce \textit{scene graphs}~\cite{MitraCCoT,LLaVA-SG}, which provide a structured representation of images by explicitly modeling visual objects and their spatial relations.

Crucially, although both lines of research adopt a similar strategy—utilizing structured representations via commonsense and scene graphs—they are typically treated as independent components.
This separation overlooks the potential synergy derived from their structural similarity and semantic complementarity.
Concretely, as illustrated in Figure~\ref{fig:motivation}, scene graphs excel at visual grounding, capturing explicit entities and their dynamic interactions (e.g., \textit{holding}, \textit{next to}) directly from the image; however, they are limited to visible content and lack the external context required for deep reasoning.
In contrast, commonsense graphs encapsulate rich semantic priors and relational knowledge (e.g., \textit{is friend of}, \textit{used for}). Yet, their effectiveness heavily relies on accurate entity alignment; without precise visual anchoring, the retrieval process is susceptible to visual noise, often introducing irrelevant or misleading information.

Motivated by these observations, we propose to bridge visual perception and external knowledge by fusing scene and commonsense graphs into a unified structured representation to empower reasoning.
While conceptually intuitive, realizing this synergy poses non-trivial challenges across the entire pipeline, ranging from graph construction and efficient retrieval to effective cross-modal fusion.

First, regarding graph construction, widely used commonsense graphs in VQA are predominantly textual, inherently limiting joint reasoning across modalities~\cite{mKG-RAG}.
Therefore, transforming them into Multi-modal Knowledge Graphs (MMKGs)~\cite{liu2019mmkg} is essential for reliable answer generation.
Nevertheless, existing MMKGs~\cite{multimodalKG, MMKG-survey} are largely generic, and pipelines for constructing domain-specific MMKGs tailored to complex VQA tasks are not yet well-established.

Second, in retrieval, bridging visual objects with external knowledge remains challenging. 
Existing VQA methods~\cite{filterRAG, llm-ra} typically rely on coarse global vision–text alignment, which often overlooks fine-grained visual cues (e.g., interactions) and cause ambiguity when multiple entities share similar textual descriptions.
Finally, regarding graph fusion, the inherent heterogeneity between commonsense and scene graphs poses a significant obstacle. Since the two graphs differ in modality, topology, and node attributes, achieving reliable node alignment and coherent cross-modal fusion remains a challenge.

To address these issues, we propose {\bf KG-ViP}, a unified framework that bridges fine-grained visual perception and external knowledge.
First, to ensure knowledge support, we introduce a robust pipeline for constructing domain-specific commonsense graphs in the form of MMKGs. This pipeline accommodates diverse data scenarios, constructing the graph from heterogeneous sources and enriching entities with visual attributes.
Building on this, we design a novel Retrieval–Fusion–Generation workflow for KG-ViP. Specifically, we employ a two-stage retrieval strategy that integrates textual and visual modalities.  It utilizes expanded query semantics derived from textual retrieval to prune the noisy scene graph, and subsequently employs the refined visual cues to conduct visual retrieval, thereby precisely anchoring external knowledge. This process yields query-aware scenes and commonsense subgraphs.
To fuse these heterogeneous graphs, we implement a comprehensive strategy comprising cross-modal entity alignment and agent-based refinement, resulting in a unified graph.
Finally, this graph is fed into the MLLMs, providing enhanced support for reliable reasoning.

The main contributions of this paper are summarized as follows: {\bf (1)} We identify the dual limitations of MLLMs in knowledge grounding and visual perception, advocating for a synergistic integration of scene and commonsense graphs to enhance joint reasoning; {\bf (2)} To realize this, we propose KG-ViP, a unified framework featuring a pipeline for graph construction, a two-stage retrieval strategy, and a cross-modal fusion mechanism; 
{\bf (3)} Extensive empirical studies demonstrate the effectiveness and superiority of our framework across various evaluation tasks.

%% file: Section/Related_Work.tex
\section{Related Work}

\subsection{MLLMs for Knowledge-based VQA}
Knowledge-Based VQA (KB-VQA) typically requires incorporating external knowledge to answer questions that demand information beyond visual content. Recently, with the rapid advancement of MLLMs, leveraging MLLMs for KB-VQA has garnered significant research attention~\cite{KAT,combo,Fine-grained-VQA}. In particular, retrieval augmentation has proven effective in enhancing KB-VQA performance, as demonstrated by representative methods such as RAVQA~\cite{RAVQA}, Wiki-LLaVA~\cite{Wiki-LLaVA}, EchoSight~\cite{EchoSight}, FilterRAG~\cite{filterRAG}, and LLM-RA~\cite{llm-ra}. Specifically, FilterRAG retrieves knowledge by jointly embedding the question and the image split into $2\times2$ patches, while LLM-RA employs MLLMs to generate image captions from which key entities are extracted for retrieval. However, these methods largely overlook the rich structural scene information and fine-grained details inherent in images. We explicitly model images as scene graphs to guide knowledge retrieval, significantly enhancing answer accuracy.

\subsection{Scene Graph for MLLM Reasoning}
Scene graphs bridge the visual-semantic gap by explicitly modeling objects and their spatial interactions~\cite{SSG-Survey,SSGsurvey}. While foundational in early VQA, they have recently become instrumental in enhancing the fine-grained perception of MLLMs~\cite{llmmeetsg,Kim_2024_CVPR}. For instance, CCoT~\cite{MitraCCoT} interprets images as scene graphs to augment CoT prompting for MLLMs. M3COT~\cite{M3COT} improves VQA robustness by unifying multi-view scene graphs, while MMCD~\cite{MMCD} employs structural perturbations of scene graphs to suppress model hallucinations.
However, these methods primarily treat scene graphs as internal visual parsing artifacts and overlook the potential role as explicit anchors for external knowledge. 
Consequently, they remain limited in knowledge-intensive reasoning, motivating the integration of scene and commonsense graphs.

%% file: Section/Method.tex
\section{Methodology}

\begin{figure*}[h!]
    \centering
    \includegraphics[width=1\textwidth]{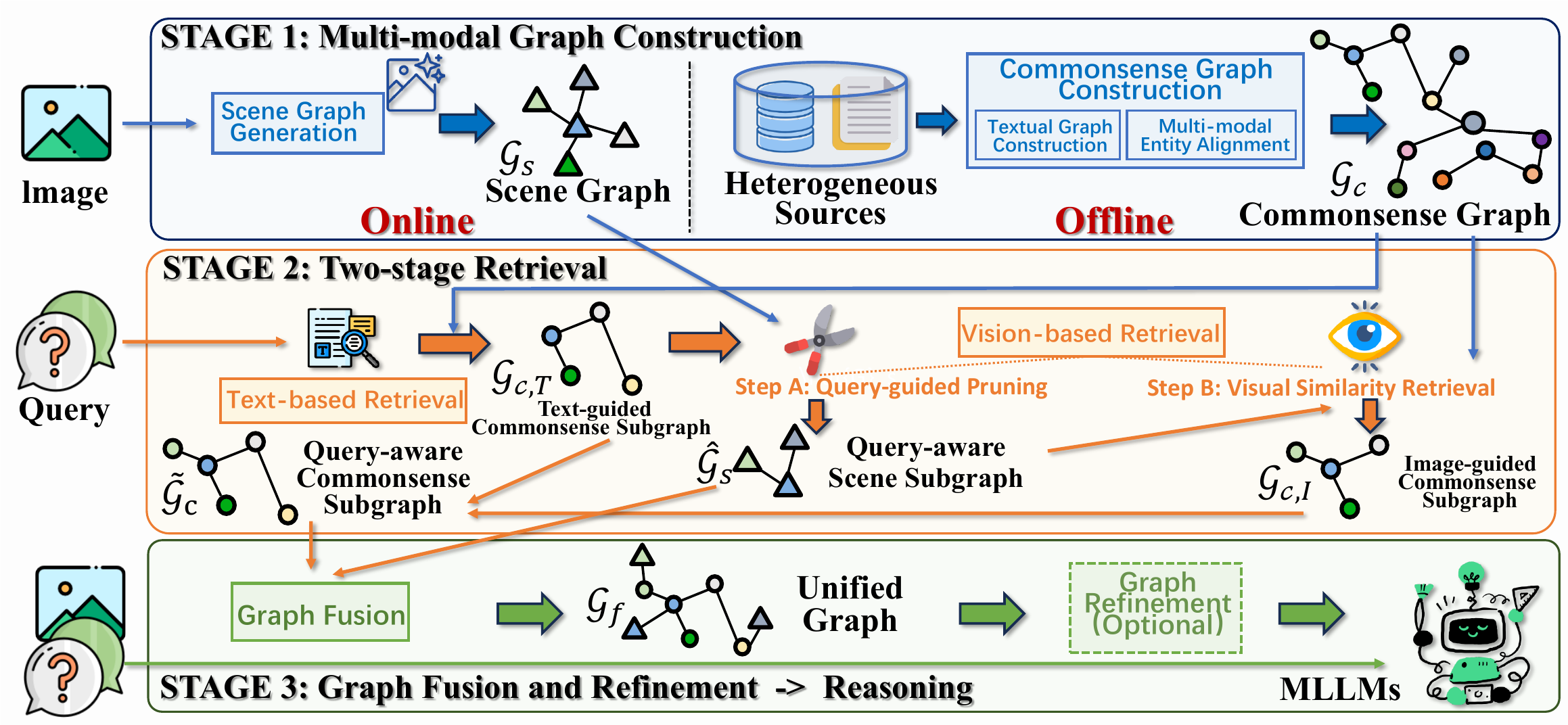}
    \vspace{-10pt}
    \caption{{\bf The Overview of KG-ViP Framework.}
    \footnotesize KG-ViP operates in three stages.
First, we prepare a domain \emph{Commonsense Graph} $\mathcal{G}_c$ and derive a \emph{Scene Graph} $\mathcal{G}_s$ from the input image $I_q$.
Second, the query $Q$ retrieves a text-guided subgraph $\mathcal{G}_{c,T}$ from $\mathcal{G}_c$, which is used to filter $\mathcal{G}_s$ into a refined $\hat{\mathcal{G}}_s$. Subsequently, $\hat{\mathcal{G}}_s$ guides the retrieval of visual knowledge $\mathcal{G}_{c,I}$ to form the final commonsense subgraph $\tilde{\mathcal{G}}_c$.
Finally, $\hat{\mathcal{G}}_s$ and $\tilde{\mathcal{G}}_c$ are fused into a unified graph $\mathcal{G}_f$, providing the structured context for the MLLM to generate the answer $A$.}
    \label{fig:framework}
    \vspace{-10pt}
\end{figure*}

In this work, we focus on Knowledge-based  VQA. Given an image $I_q$ and a textual query $Q$, the goal is to generate an answer $A$ by bridging visual perception with external knowledge. As illustrated in Figure~\ref{fig:framework}, our KG-ViP framework comprises three distinct stages, detailed as follows.

\subsection{Multi-modal Graph Construction}
\label{sec:graph_construction}

An MMKG extends traditional KGs by integrating structured relational facts with heterogeneous data from diverse modalities. Formally, we define an MMKG as $\mathcal{G} = (\mathcal{E}, \mathcal{R})$, where $\mathcal{E}$ denotes the set of entities (nodes), and $\mathcal{R}$ represents the set of relations (edges) connecting these entities. Each entity $e \in \mathcal{E}$ is associated with various attributes, encompassing multi-modal content (e.g., images) and metadata (e.g., textual labels). In this work, we construct two distinct types of MMKGs to facilitate visual reasoning: \textbf{(1) Commonsense Graph $\mathcal{G}_c$:} An MMKG encapsulating large-scale, domain-specific commonsense facts spanning diverse modalities. For instance, within the cinematic domain, this graph organizes character relationships and plot synopses into an MMKG structure, enriched with visual content such as character portraits.
\textbf{(2) Scene Graph $\mathcal{G}_s$:} An MMKG derived from the specific query image provided for MLLM reasoning tasks. It captures concrete objects along with their spatial relationships and interactions to aid query-specific visual understanding. For example, given a single movie still, this graph explicitly models the spatial configurations and actions of the depicted characters and objects.

\subsubsection{Commonsense Graph Construction}
We prioritize using off-the-shelf commonsense graphs when available for the target domain. Otherwise, we construct the graph from heterogeneous sources (e.g, multi-modal data, existing textual knowledge bases) via a two-step pipeline: \emph{Textual Graph Construction} and \emph{Multi-modal Entity Alignment}. Notably, the pipeline is orthogonal to specific MMKG construction techniques, allowing for the integration of various existing methods~\cite{MMKG-survey,multimodalKG,mKG-RAG}. We outline the general workflow below.

\noindent{\bf Textual Graph Construction.} The first stage transforms heterogeneous sources—including unstructured texts and existing structured knowledge bases—into a unified graph. Crucially, we leverage LLMs to extract salient entities and explicit relations from unstructured texts (e.g., movie plot synopses)~\cite{graphrag,lightrag}, constructing the graph from scratch. Additionally, we can integrate existing structured knowledge bases by consolidating independent triples via standard techniques, such as embedding-based entity alignment~\cite{OpenEA} and semantic relation normalization~\cite{Relation-emb,llmextractkg}. To ensure topological quality, the graph undergoes refinement—such as pruning low-confidence edges and extracting the giant connected component~\cite{KGPruning}.

\noindent{\bf Multi-modal Entity Alignment} The second stage enriches the textual entities with corresponding multi-modal objects (e.g., images or video clips).  For entities associated with official media repositories or explicit metadata, we employ high-precision matching based on semantic consistency. Conversely, for entities lacking direct associations, we utilize an LLM-driven web search agent~\cite{MultiReflect,DEFAME} to retrieve and filter candidate objects. For instance, to visually ground the entity \textit{``Iron Man''}, the LLM agent retrieves a set of movie stills and retains only those where visual attributes (e.g., red and gold armor) statistically align with the entity's textual definition, selecting the top-$k$ most representative instances.

\subsubsection{Scene Graph Generation}
For a specific query, we generate a scene graph that transforms the raw image into explicit objects along with their spatial relationships and interactions. 
Given an input image $I_q$, we employ an MLLM or a specialized scene graph generation model~\cite{EGTR,ACC,Open-World-SGG,Kim_2024_CVPR} to detect a set of objects $\mathcal{E}_s$ and identify the relations $\mathcal{R}_s$ between pairs of objects. Notably, to prevent the scene graph from merely replicating the MLLM's inherent coarse-grained perception, we can employ optimization strategies such as region-wise scanning~\cite{GPT4SGG}, multi-turn iterative refinement~\cite{ISGG}, and ensemble with specialized models~\cite{EGTR,Grounded-SAM} to extract dense visual details. Consequently, this process yields the scene graph $\mathcal{G}_s = (\mathcal{E}_s, \mathcal{R}_s)$. 
Crucially, the detected objects $\mathcal{E}_s$ serve as both fine-grained visual grounding and structural anchors for external knowledge retrieval.

\subsection{Two-stage Retrieval}

Since large-scale commonsense graphs typically contain millions of relational facts, efficiently retrieving information relevant to the query is critical. 
Existing methods~\cite{EVQACK,Fine-grained-VQA,mmgraphrag} typically rely on text-text or vision-text embedding matching for retrieval, which overlooks vision-centric retrieval and suffers from two limitations: (1) \textit{Textual Bias}: relevant knowledge may be missed if the key visual entities are not explicitly mentioned in the textual query; 
(2) \textit{Visual Noise}: vision–text matching may incorrectly retrieve semantically related but contextually irrelevant nodes, such as different characters played by the same actor or different items within the same category. 
To address these issues, we introduce a two-stage retrieval method that comprehensively extracts query-relevant commonsense information from both text and visual modalities.

{\bf Stage 1: Text-based Retrieval.}
The first stage focuses on acquiring a broad semantic context derived from the textual query.
Given a commonsense graph $\mathcal{G}_c = (\mathcal{E}_c, \mathcal{R}_c)$ and a query $Q$, we identify a set of entities $\mathcal{E}_{text}$ from $Q$. For each entity $t \in \mathcal{E}_{text}$, we utilize the text embedding similarity to find the corresponding node in $\mathcal{G}_c$ and extract a $k$-th order subgraph centered on it. 
This extraction significantly reduces the search space to ensure scalability.
The union of these subgraphs forms an initial candidate subgraph $\mathcal{G}_{c}^k = (\mathcal{E}_{c}^k, \mathcal{R}_{c}^k) \subseteq \mathcal{G}_c$.

To further remove irrelevant or weakly related entities, we estimate the importance of each entity $e\in \mathcal{E}_{c}^k$ by a generalized ranking mechanism, such as Random Walk with Restart (RWR) ~\cite{RWR} or Personalized PageRank (PPR) ~\cite{PPR}.
We then select the top-$n$ ranked entities $\mathcal{E}_{c,T} \subseteq \mathcal{E}_{c}^k$, together with the relations among them $\mathcal{R}_{c,T} \subseteq \mathcal{R}_c^k$, to form a compact text-guided commonsense subgraph $\mathcal{G}_{c,T} = (\mathcal{E}_{c,T}, \mathcal{R}_{c,T})$.

{\bf Stage 2: Vision-based Retrieval.}
Visual input $I_q$ provides complementary entity cues for VQA that are often absent from the textual query $Q$. However, naively aligning all entities from the scene graph $\mathcal{G}_s$ derived from $I_q$ with the commonsense graph is inefficient. Owing to the long-tailed distribution of scene graph entities, many detected objects are weakly related or irrelevant to the query, leading to high retrieval cost and noisy knowledge grounding. To overcome this, we propose a coarse-to-fine strategy involving \emph{Query-guided Pruning} followed by \emph{Visual Similarity Retrieval}.

\noindent{\underline{ \it Step A: Query-guided Pruning.}}
We first utilize the text-guided commonsense subgraph $\mathcal{G}_{c,T}$ obtained in Stage 1 to filter visual noise. Specifically, we prompt an MLLM to prune the original scene graph $\mathcal{G}_s$ by removing entities that are irrelevant to the $\mathcal{G}_{c,T}$.
The remaining entities $\hat{\mathcal{E}}_s$ and their relations form a query-aware scene subgraph $\hat{\mathcal{G}}_s = (\hat{\mathcal{E}}_s, \hat{\mathcal{R}}_s)$. 

This strategy offers significant advantages over traditional query-based filtering.
While existing VQA approaches~\cite{RAVQA,llm-ra} typically identify salient visual entities by conditioning on the textual query $Q$, such filtering is limited by the sparse semantic scope of the query and may discard entities that are implicitly relevant through relational dependencies.
For example, a visual entity not mentioned in the query may still be related to the queried concepts via commonsense relations (e.g., a tool associated with a queried action).
By conditioning on $\mathcal{G}_{c,T}$, which encodes a richer set of entities and local relations, we enable more comprehensive entity recognition from redundant visual information.

Moreover, pruning the scene graph with respect to $\mathcal{G}_{c,T}$ facilitates subsequent graph fusion (See Section~\ref{sec:fusion_refinement}). Restricting $\hat{\mathcal{G}}_s$ to a semantic subspace overlapping with the commonsense graph reduces structural incompatibility and alignment ambiguity, thereby enabling more reliable cross-modal entity alignment and knowledge integration.

\noindent{\underline{\it Step B: Visual Similarity Retrieval.}}
With the refined visual entities $\hat{\mathcal{E}}_s \in \hat{\mathcal{G}}_s$, we proceed to retrieve complementary knowledge from the commonsense graph $\mathcal{G}_c$.
Prior knowledge-based methods~\cite{infoseek, filterRAG} commonly retrieve relevant commonsense graph entities by computing vision–to-text embedding similarity, treating textual descriptions as proxies for visual objects.
However, such alignment may overlook subtle visual cues and cause ambiguity when multiple entities share similar textual descriptions.

In contrast, we perform direct vision-to-vision retrieval by matching the visual regions of scene graph entities $e_s \in \hat{\mathcal{E}}_s$ with the visual attributes (i.e., representative images) of candidate entities $e_c \in \mathcal{E}_c$ in the commonsense graph. 
Specifically, we extract embeddings for $e_s$ and $e_c$ using vision encoders (e.g., Cambrian-1~\cite{cambrian1}, CLIP~\cite{CLIP}) and compute similarity scores between each pair. 
A generalized ranking mechanism is then applied to select the top-$m$ most relevant entities $\mathcal{E}_{c,I} \subseteq \mathcal{E}_c$, forming the vision-guided commonsense subgraph $\mathcal{G}_{c,I} = (\mathcal{E}_{c,I}, \mathcal{R}_{c,I})$.

Finally, we take the union of the text-guided subgraph $\mathcal{G}_{c,T}$ and the vision-guided subgraph $\mathcal{G}_{c,I}$ to form a unified query-aware commonsense subgraph:
$
\tilde{\mathcal{G}}_c = \mathcal{G}_{c,T} \cup \mathcal{G}_{c,I} = (\tilde{\mathcal{E}}_c, \tilde{\mathcal{R}}_c) \subseteq {\mathcal{G}}_c$,
which aggregates complementary knowledge retrieved from both modalities and provides the necessary context for the VQA task.

\subsection{Graph Fusion and Refinement}
\label{sec:fusion_refinement}

We have obtained two query-aware heterogeneous subgraphs: the scene subgraph $\hat{\mathcal{G}}_s$ that provides fine-grained visual details, and the commonsense subgraph $\tilde{\mathcal{G}}_c$ that offers rich knowledge grounding. 
Maintaining them as a separate structure risks a semantic disconnect. 
For example, MLLMs may fail to anchor a generic scene node like "man holding a cup" in $\hat{\mathcal{G}}_s$ to the specific identity of "Detective Cobb" in $\tilde{\mathcal{G}}_c$. 
Such fragmented representations hinder the synthesis of visual evidence and external knowledge, and may even introduce spurious or misleading information during reasoning.

To address this issue, we fuse the two subgraphs into a unified representation by explicitly aligning nodes that refer to the same real-world entity.
Since textual labels in the scene graph typically describe generic visual attributes (e.g., appearance or common object categories) rather than precise identities, direct text-to-text matching is unreliable. We therefore prioritize visual grounding for cross-graph alignment.
Specifically, for each pair $(e_i, e_j) \in \hat{\mathcal{E}}_s \times \tilde{\mathcal{E}}_c$, we compute a  similarity $S_{ij}$ :
{
\setlength{\abovedisplayskip}{3pt} 
\setlength{\belowdisplayskip}{3pt}
\begin{equation*}
S_{ij} = \alpha \langle \phi(I_i), \psi(T_j) \rangle + (1 - \alpha) \langle \phi(I_i), \phi(I_j) \rangle,
\end{equation*}
}
where $\phi$ and $\psi$ denote the visual and text encoders, respectively, $ I_i$ is the cropped visual region of the scene node $e_i$; $T_j$ and $I_j$ denote the textual content and the associated image of the commonsense node $e_j$.
The first term measures the cross-modal alignment between the visual object and the knowledge concept, while the second term captures the visual similarity between the object and the entity's visual reference.
For aligned entity pairs, we merge their attributes to construct a coherent representation: we prioritize visual properties from $\hat{\mathcal{G}}_s$ to ensure strict grounding in the image context, while augmenting them with fine-grained description and labels from $\tilde{\mathcal{G}}_c$ to enhance semantic precision. The resulting unified graph $\mathcal{G}_f$ encapsulates both the dynamic visual details and the external knowledge.

To mitigate potential redundancy or missing links in the consolidated graph $\mathcal{G}_f$, we employ an optional MLLM-driven agent~\cite{selfrag,SAMRAG,MIRAG} to iteratively refine the topology by evaluating its alignment with $Q$. At each step, the agent selects one of three actions: (1) \textbf{Expand}, to retrieve missing critical nodes that bridge disconnected evidence; (2) \textbf{Prune}, to remove noise or irrelevant entities; and (3) \textbf{Terminate}, to conclude the process. This refinement operates for a maximum of $t$ steps, ensuring $\mathcal{G}_f$ becomes structurally concise and semantically aligned with the user's intent before final reasoning. 
Finally, we input $\mathcal{G}_f$ into the MLLM, together with the original visual input $I_q$ and the textual query $Q$, to produce more reliable responses.

%% file: Section/Experiment.tex
\section{Evaluation}

\input{Table/VQA_result}

\subsection{Benchmarks}
\label{sec:benchmarks}

We evaluate our framework on two benchmarks: 

\noindent\textbf{FVQA 2.0+.} 
First, we adopt FVQA 2.0~\cite{FVQA2-0}, a widely used benchmark for VQA requiring external knowledge. Since its original knowledge base consists solely of textual triplets, we upgrade it to FVQA 2.0+ to fit the multi-modal setting. Following the \emph{Commonsense Graph Construction} pipeline in Section~\ref{sec:graph_construction}, we enrich the original entities with visual representations, resulting in a commonsense graph with 1,152 nodes, 1,767 edges, and 3,342 images. We utilize the original 2,820 QA samples, where each entry comprises a query image, a question requiring external knowledge, and a ground-truth natural language answer.

\noindent\textbf{MVQA.} 
Since FVQA 2.0 primarily involves simple one-hop reasoning over the knowledge graph, we introduce a new benchmark, MVQA, designed to evaluate complex multi-hop reasoning. Constructed based on 50 movies from MovieBench~\cite{moviebench}, we built a cinematic commonsense graph using the same construction methodology, comprising 1,271 nodes, 1,468 edges, and 4,611 images. Furthermore, we developed an automated generation pipeline (detailed in Appendix~\ref{Benchmark Detail}) to create 1,433 complex QA samples. Unlike previous datasets, these queries necessitate deep reasoning that bridges fine-grained visual cues in the query image with multi-hop relational paths in the commonsense graph.

\subsection{Experimental Setup}

\noindent{\bf Baselines.}
We compare \textbf{KG-ViP} against three categories of baselines:
(i) \emph{vanilla MLLM} methods that rely on the intrinsic reasoning ability of MLLMs, including \textbf{Zero-shot}, \textbf{Fine-tuning}, and \textbf{Chain-of-Thought (CoT)}~\cite{COT};
(ii) \emph{scene graph–augmented} methods, represented by \textbf{CCoT}~\cite{MitraCCoT}, which converts images into scene graphs to enhance CoT reasoning;
(iii) \emph{retrieval-augmented} methods that incorporate external knowledge, including \textbf{FilterRAG}~\cite{filterRAG} and \textbf{LLM-RA}~\cite{llm-ra}.
In addition, we implement a RAG baseline using a naive text-based retrieval strategy, termed \textbf{NaiveRAG}.
Implementation details are provided in Appendix~\ref{Experiment Detail}.

\noindent{\bf Evaluation Metrics.}
Following prior work~\cite{MMBench,MMed-RAG, WearVQA}, we adopt three widely used metrics for evaluation.
METEOR~\cite{METEOR} calculates the harmonic mean of unigram precision and recall, capturing semantic variations while applying a fragmentation penalty for word order.
Semantic Answer Similarity (SAS)~\cite{SAS} utilizes a cross-encoder to assess the semantic equivalence between predicted and reference answers.
LLM-as-a-Judge~\cite{LLM-as-a-Judge} leverages an LLM to assign a scalar score that evaluates the relevance of the generated response with the ground-truth answer.
We denote this metric as LLM-J for short.

\noindent{\bf Implementation Details.}
We adopt Qwen2.5-VL-7B as the backbone MLLM for entity extraction, scene graph generation, graph refinement, and final answer generation.
For retrieval, all-MiniLM-L6-v2 and Cambrian-1-8B are employed as the text and visual encoders, respectively.
The iteration of graph refinement is fixed to \(t=1\) to balance refinement quality and efficiency.
Qwen2.5-7B and DeepSeek-V3.2 (671B) are used as judge models for the LLM-J metric.
All experiments are conducted on four NVIDIA A100 GPUs.
More details are provided in Appendix~\ref{Experiment Detail}
\subsection{Main Results}

Table~\ref{tab:vqa_comparison_results} presents the main results on FVQA 2.0+ and MVQA.
Overall, our proposed KG-ViP consistently achieves the best performance across all evaluation metrics on both benchmarks, showing its strong multi-modal reasoning capability.
Specifically, on FVQA 2.0+, KG-ViP improves the LLM-J (DeepSeek-V3.2) by 7.76\% and the SAS by 2.93\%, compared to the previous leading baseline LLM-RA. 
The results underscore the importance of incorporating structured representations for visual inputs in VQA.
While FilterRAG and LLM-RA also identify key visual entities and retrieve related external knowledge, they largely ignore the spatial relations among visual entities. In contrast, KG-ViP explicitly models entities and their spatial interactions with the scene graph, thereby improving the visual perception and grounding of MLLMs. 
MVQA poses a more challenging setting that involves complex multi-hop reasoning over multiple entities and relations, demanding precise alignment between visual grounding and external knowledge. 
In this scenario, KG-ViP exhibits an even larger performance margin, outperforming LLM-RA by 11.34\% in the LLM-J (DeepSeek-V3.2) and 6.03\% in SAS. 
We attribute these gains to the unified and compact graph representation produced by KG-ViP, which integrates the scene graph and commonsense graph, providing a low-noise and structured context to enhance multi-modal reasoning.

\input{Table/VQA_abt_1}

\input{Table/VQA_abt_2}

\input{Table/VQA_abt_3}

\input{Table/Different_model}

\begin{figure}[h]
    \centering
    \makebox[0pt][l]{\hspace*{-3.8cm}\includegraphics[width=0.47\textwidth]{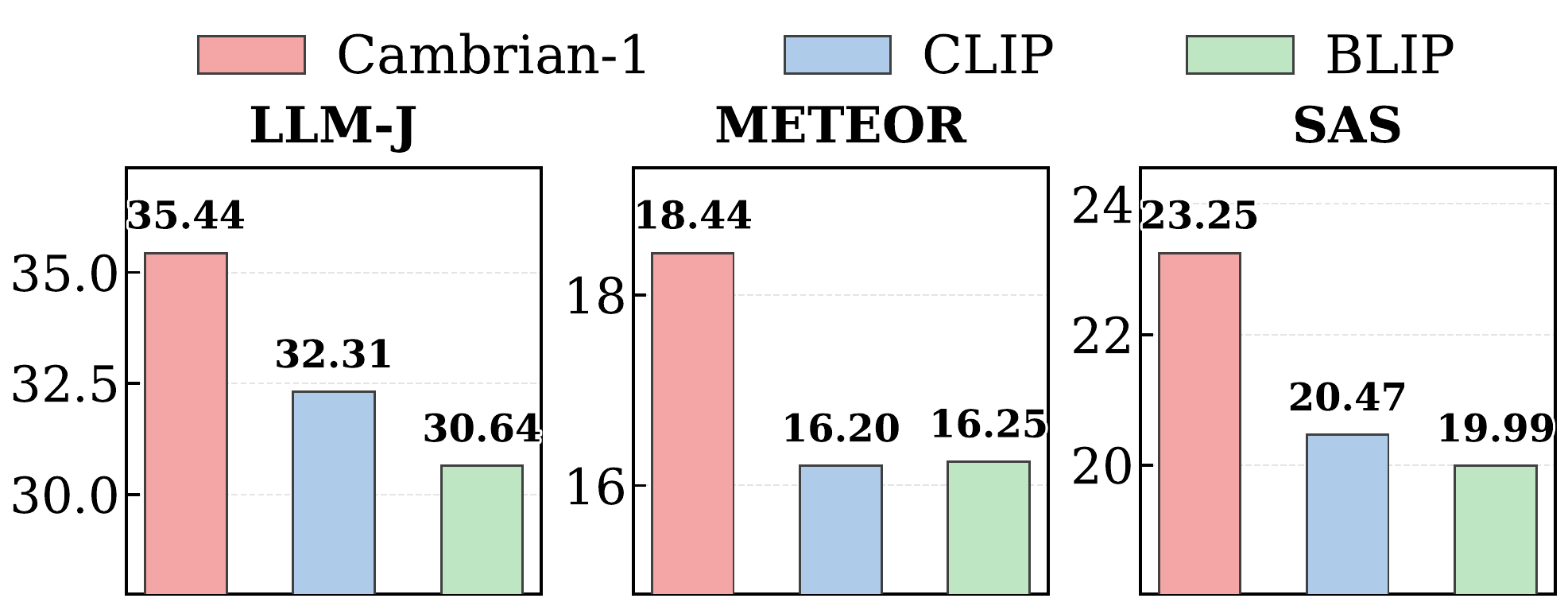}}
    \vspace{-5pt}
\caption{Impact of vision encoder selection on KG-ViP.}
    \label{fig:com_dif_emb_model}
    \vspace{-15pt}
\end{figure}

\subsection{Ablations and Analysis}
We perform a comprehensive ablation and analysis study of KG-ViP.
For the LLM-J metric, DeepSeek-V3.2 serves as the default judge model.

\noindent\textbf{Effectiveness of Graph Fusion and Refinement.}
To validate the effectiveness of graph fusion and refinement, we evaluate two ablated variants: (i) providing the MLLM with isolated scene and commonsense subgraphs without fusion, and (ii) using the fused graph without refinement.
As shown in Table~2, removing graph fusion results in substantial performance degradation on both benchmarks, with the LLM-J dropping by 5.64\% on MVQA and 2.29\% on FVQA 2.0+.
This highlights the critical role of a unified graph representation in synergizing visual evidence with commonsense knowledge.
Furthermore, disabling the graph refinement step also leads to a noticeable decline in performance, demonstrating that pruning redundant nodes is essential for maintaining a high-quality, task-relevant graph structure.

\noindent\textbf{Impact of Retrieval Mode.}
KG-ViP leverages both textual and visual modalities to extract query-relevant commonsense information.
Regarding the visual branch, retrieval can be implemented via either vision-to-text (V$\to$T) or vision-to-vision (V$\to$V) matching.
KG-ViP specifically adopts V$\to$V matching to mitigate the ambiguity inherent in V$\to$T approaches, where distinct visual entities may share similar textual descriptions.
We evaluate the performance of these retrieval configurations in Table~\ref{table:retrieval_mode}.
The results indicate that the absence of text-based retrieval leads to consistent performance degradation regardless of the vision-based mode, confirming that both modalities provide indispensable cues.
Furthermore, when text-based retrieval is retained, V$\to$V matching consistently outperforms V$\to$T matching.
These findings suggest that direct visual alignment captures subtle visual cues more effectively, leading to more reliable knowledge retrieval and reasoning.

\noindent\textbf{Impact of Scene Graph Pruning.}
To mitigate visual noise stemming from the long-tailed distribution of scene graphs, KG-ViP prunes task-irrelevant entities by conditioning the scene graph on the text-guided commonsense subgraph. To validate this strategy, we compare it against Query Pruning, which relies solely on the raw query text for filtering. 
As shown in Table 4, our proposed graph pruning consistently outperforms both no pruning and Query Pruning.
This demonstrates that while pruning the scene is essential for noise reduction, leveraging the enriched entity set from the commonsense subgraph provides a more robust semantic reference than the sparse query text, effectively preserving critical visual evidence for reasoning.

\noindent\textbf{Selection of Vision Encoders.}
To analyze the impact of visual representation on KG-ViP, we investigate several vision encoders, including CLIP, BLIP, and Cambrian-1, on the MVQA benchmark. 
As illustrated in Figure~\ref{fig:com_dif_emb_model}, Cambrian-1 significantly outperforms both CLIP and BLIP. 
We attribute this performance gap to the distinct nature of these models: CLIP and BLIP are primarily designed as text-vision alignment models, which prioritize matching global visual features with textual descriptions. 
In contrast, Cambrian-1 functions as a specialized vision-centric encoder, optimized for capturing fine-grained visual details and spatial structures.
These advantages make Cambrian-1 particularly well-suited for the vision-centric retrieval and reasoning scenarios required by KG-ViP.

\noindent\textbf{Robustness across MLLM Architectures.}
Table~\ref{table:mllm_comparison} presents a comprehensive comparison across MLLMs of varying scales on MVQA, including Qwen2.5-VL-3B/32B and GLM-4.1V-9B/GLM-4.5V-106B. When integrated with our KG-ViP framework, these models exhibit consistent performance gains, with average improvements of 20.23\%, 7.85\%, and 15.06\% across the LLM-J, METEOR, and SAS metrics, respectively. This confirms KG-ViP's robustness across diverse MLLM architectures and scales.

%% file: Table/VQA_result.tex
\begin{table*}[t]
\centering
\renewcommand{\arraystretch}{0.95} 
\setlength{\tabcolsep}{3pt}

\caption{
\textbf{Comparison of different methods on \textsc{FVQA 2.0+} and \textsc{MVQA}.}
\footnotesize
Results are reported in terms of LLM-J (LLM-as-a-Judge), METEOR, and SAS.}
\label{tab:vqa_comparison_results}
\vspace{-0.3cm}

\resizebox{\textwidth}{!}{%
\begin{tabular}{l cccc cccc}
\toprule
\multirow{3.5}{*}{\textbf{Method}} & \multicolumn{4}{c}{\textbf{FVQA 2.0+}} & \multicolumn{4}{c}{\textbf{MVQA}} \\
\cmidrule(lr){2-5} \cmidrule(lr){6-9}
 & \multicolumn{2}{c}{LLM-J $\uparrow$} & \multirow{2}{*}{METEOR$\uparrow$} & \multirow{2}{*}{SAS$\uparrow$}
 & \multicolumn{2}{c}{LLM-J $\uparrow$} & \multirow{2}{*}{METEOR$\uparrow$} & \multirow{2}{*}{SAS$\uparrow$} \\
\cmidrule(lr){2-3} \cmidrule(lr){6-7}
 & Qwen2.5-7B & DeepSeek-V3.2 & & & Qwen2.5-7B & DeepSeek-V3.2 & & \\
\midrule

\cellcolor[gray]{0.95} & \multicolumn{8}{c}{\cellcolor[gray]{0.95} \textit{Vanilla MLLM Reasoning}} \\
Zero-shot                 & 36.79 & 38.90 & 15.12 & 32.70 & 14.41 & 15.84 & 7.70  & 9.18  \\
CoT           & 39.96 & 40.35 & 18.30 & 36.65 & 16.09 & 17.78 & 8.20  & 9.69  \\

% Caption-Based           & 41.03 & xx & 16.63 & 35.84 & 16.94 & xx & 8.91  & 9.79  \\
% Few-shot           & 42.71 & xx & 19.18 & 40.65 & 16.32 & xx & 8.46  & 10.81  \\

Fine-Tuning               & 43.16 & 43.09 & 19.82 & 39.52 & 16.23 & 16.85 & 8.33  & 11.22 \\

\cellcolor[gray]{0.95} & \multicolumn{8}{c}{\cellcolor[gray]{0.95} \textit{Scene Graph–Augmented Reasoning}} \\
% Capstone\cite{CAPSTONE}      & 41.28 & xx & 17.37 & 37.01 & 16.04 & xx & 8.08  & 10.21 \\
CCoT\cite{MitraCCoT}      & 43.34 & 44.32 & 20.35 & 40.56 & 16.41 & 18.64 & 8.69  & 10.23 \\
% PFVR\cite{PFVR}      & 45.53 & xx & 17.29 & 38.50 & 17.34 & xx & 8.34  & 10.22 \\

\cellcolor[gray]{0.95} & \multicolumn{8}{c}{\cellcolor[gray]{0.95}\textit{Retrieval-Augmented Reasoning}} \\
NaiveRAG                  & 46.25 & 48.09 & 21.52 & 43.93 & 19.47 & 20.78 & 9.11  & 13.17 \\
FilterRAG\cite{filterRAG} & 50.16 & 51.73 & 22.46 & 45.72 & 20.11 & 20.84 & 12.41 & 15.49 \\
LLM-RA \cite{llm-ra}      & 51.90 & 52.64 & 22.96 & 46.82 & 22.53 & 24.10 & 11.43 & 17.22 \\
% MMGraphRAG\cite{MMGraphRAG-B}      & 52.86 & xx & 23.44 & 47.24 & 27.93 & xx & 17.33  & 18.51 \\
% MegaRAG\cite{MegaRAG}      & 50.92 & xx  & 23.72 & 47.25 & 28.32 & xx  & 17.56& 18.82 \\
\rowcolor{rowblue}
\textbf{KG-ViP (Ours)}    & \textbf{59.50} & \textbf{60.40} & \textbf{23.83} & \textbf{50.75}
                          & \textbf{30.46} & \textbf{35.44} & \textbf{18.44} & \textbf{23.25} \\
\bottomrule
\end{tabular}%
}
\vspace{-0.4cm} 
\end{table*}

%% file: Table/VQA_abt_1.tex
\newcommand{\drop}[1]{\textcolor{green!70!black}{$-$#1}}

\begin{table}[t]
\centering
\caption{
Effectiveness of Graph Fusion and Refinement
}
\label{table:abs_vqa}
\vspace{-5pt}
\setlength{\tabcolsep}{4pt} 
\renewcommand{\arraystretch}{0.9}

\resizebox{0.48\textwidth}{!}{%
\begin{tabular}{ccccc}
\toprule
\multirow{2}{*}{\textbf{Method}} & \multicolumn{2}{c}{\textbf{FVQA 2.0+}} & \multicolumn{2}{c}{\textbf{MVQA}} \\
\cmidrule(lr){2-3} \cmidrule(lr){4-5}

 & LLM-J$\uparrow$ & METEOR$\uparrow$
 & LLM-J$\uparrow$ & METEOR$\uparrow$ \\

\midrule

\rowcolor{rowblue}
\textbf{KG-ViP}
& \textbf{60.40} & \textbf{23.83} 
& \textbf{35.44} & \textbf{18.44} \\

\addlinespace[2pt]

w/o Graph Fusion
& 58.11 & 23.43 
& 29.80 & 15.10 \\
$\Delta$
& \drop{2.29} & \drop{0.40} 
& \drop{5.64} & \drop{3.34} \\
\addlinespace[2pt]

w/o Graph Refinement
& 60.18 & 23.52 
& 33.19 & 18.19 \\
$\Delta$
& \drop{0.22} & \drop{0.31} 
& \drop{2.25} & \drop{0.25} \\

\bottomrule
\end{tabular}%
}
\end{table}

%% file: Table/VQA_abt_2.tex
\newcommand{\yes}{\ding{51}}

\begin{table}[t]
\vspace{-5pt}
\centering
\caption{
Impact of Retrieval Mode
}
\label{table:retrieval_mode}
\vspace{-5pt}

\setlength{\tabcolsep}{4pt} 

\renewcommand{\arraystretch}{0.9}

\resizebox{0.47\textwidth}{!}{%
\begin{tabular}{ccc cccc}
\toprule

\multirow{2}{*}{\shortstack{\textbf{Text Retrieval}}} & 
\multicolumn{2}{c}{\shortstack{\textbf{Vision Retrieval}}} &

\multicolumn{2}{c}{\raisebox{3.5pt}{\textbf{FVQA 2.0+}}} & 
\multicolumn{2}{c}{\raisebox{3.5pt}{\textbf{MVQA}}} \\
\cmidrule(lr){2-3} \cmidrule(lr){4-5} \cmidrule(lr){6-7}

 & 
\textbf{V$\to$T} & \textbf{V$\to$V} & 
LLM-J$\uparrow$ & METEOR$\uparrow$ & 
LLM-J$\uparrow$ & METEOR$\uparrow$ \\
\midrule

 & & \yes
& 58.34 & 22.66 
& 27.56 & 14.20 \\

\addlinespace[2pt]

\yes & \yes & 
& 58.66 & 23.36 
& 32.02 & 15.50 \\

\addlinespace[2pt]

\rowcolor{rowblue}
\yes & & \yes
& \textbf{60.40} & \textbf{23.83} 
& \textbf{35.44} & \textbf{18.44} \\

\bottomrule
\end{tabular}%
}
\vspace{-5pt}

\end{table}

%% file: Table/VQA_abt_3.tex
\begin{table}[t]
\centering
\caption{
Impact of Scene Graph Pruning.
}
\label{table:Scene-Graph-Pruning-Final}
\vspace{-7pt}

\setlength{\tabcolsep}{4pt} 

\renewcommand{\arraystretch}{0.9}

\resizebox{0.47\textwidth}{!}{%
\begin{tabular}{cc ccc}
\toprule
\multirow{3}{*}{\textbf{Dataset}} & \multirow{3}{*}{\textbf{Metric}} & \multicolumn{3}{c}{\textbf{Method}} \\

\cmidrule(lr){3-5}
& & \shortstack{KG-ViP \\ w/o pruning} & \shortstack{KG-ViP \\ query pruning} & \shortstack{\textbf{KG-ViP} \\ \textbf{graph pruning}} \\
\midrule

\multirow{2}{*}{\textbf{FVQA 2.0+}} 
 & LLM-J$\uparrow$ 
 & 60.04 & 59.77 & \cellcolor{rowblue}\textbf{60.40} \\

 \cmidrule(lr){2-5} 
 
 & METEOR$\uparrow$ 
 & 22.71 & 23.47 & \cellcolor{rowblue}\textbf{23.83} \\
\midrule 

\multirow{2}{*}{\textbf{MVQA}} 
 & LLM-J$\uparrow$ 
 & 30.99 & 33.39 & \cellcolor{rowblue}\textbf{35.44} \\
 \cmidrule(lr){2-5}

 & METEOR$\uparrow$ 
 & 15.88 & 17.06 & \cellcolor{rowblue}\textbf{18.44} \\

\bottomrule
\end{tabular}%
}
\end{table}

%% file: Table/Different_model.tex
\begin{table}[h!]
\centering
\caption{Robustness across  MLLM architectures.}
\label{table:mllm_comparison}
\vspace{-7pt}
\setlength{\tabcolsep}{2pt}
\renewcommand{\arraystretch}{0.9} 
\newcommand{\inc}[1]{\textcolor{red}{+#1}}

\resizebox{0.95\linewidth}{!}{%
    \scriptsize  
    \begin{tabular}{ccccc}
    \toprule
    \textbf{MLLM} & \textbf{Method} & \textbf{LLM-J$\uparrow$} & \textbf{METEOR$\uparrow$} & \textbf{SAS$\uparrow$} \\
    \midrule

    \multirow{3}{*}{\shortstack[c]{Qwen2.5-VL\\3B-Instruct}} 
        & Zero-Shot & 7.98 & 3.14 & 4.38 \\
        & \textbf{KG-ViP} & \textbf{24.39} & \textbf{5.80} & \textbf{12.51} \\
        & $\Delta$ & \inc{16.41}    & \inc{2.66}   & \inc{8.13} \\
    \midrule

    \multirow{3}{*}{\shortstack[c]{Qwen2.5-VL\\32B-Instruct}} 
        & Zero-Shot & 13.16 & 7.41 & 8.73 \\
        & \textbf{KG-ViP} & \textbf{37.35} & \textbf{18.83} & \textbf{26.96} \\
        & $\Delta$        & \inc{24.19}    & \inc{11.42}   & \inc{18.23} \\
    \midrule

    \multirow{3}{*}{\shortstack[c]{GLM-4.1V\\9B-Thinking}} 
        & Zero-Shot & 9.70 & 3.84 & 6.04 \\
        & \textbf{KG-ViP} & \textbf{28.54} & \textbf{9.50} & \textbf{19.92} \\
        & $\Delta$        & \inc{18.84}    & \inc{5.66}    & \inc{13.88} \\
    \midrule

    \multirow{3}{*}{\shortstack[c]{GLM-4.5V\\106B}} 
        & Zero-Shot & 19.10 & 10.67 & 11.70 \\
        & \textbf{KG-ViP} & \textbf{40.58} & \textbf{22.34} & \textbf{31.68} \\
        & $\Delta$        & \inc{21.48}    & \inc{11.67}   & \inc{19.98} \\
    \bottomrule
    \end{tabular}%
}
\vspace{-10pt}
\end{table}

%% file: Section/Conclusions.tex
\section{Conclusion}
In this paper, we address the dual limitations of MLLMs in knowledge grounding and visual perception for VQA. We identify that these gaps can be effectively bridged by integrating commonsense graphs, which provide external knowledge, with scene graphs, which capture fine-grained visual details. Unlike prior approaches that utilize these graphs in isolation, we propose KG-ViP, a unified framework that synergizes them through a novel retrieval-and-fusion pipeline, offering a unified structured context for joint reasoning. Extensive experiments show that our method significantly enhances performance, establishing a robust paradigm for reliable multi-modal reasoning.

%% file: Section/limitations.tex
\section*{Limitations}
While KG-ViP demonstrates robust performance in multi-modal reasoning, we identify two aspects for future exploration. 
First, our framework relies on off-the-shelf models for scene graph generation. Consequently, the quality of visual parsing sets an upper bound on downstream reasoning, particularly in challenging scenarios involving severe occlusion or small-scale objects. However, since KG-ViP is model-agnostic, it can directly benefit from future advancements in scene graph generation techniques without structural changes.
Second, to ensure online inference efficiency, we currently employ an offline-constructed commonsense graph. While this design choice minimizes latency, applying the framework to domains with evolving information would necessitate periodic graph updates or re-indexing to maintain knowledge currency. Future work could explore dynamic graph mechanisms to address such temporal shifts.

%% file: Section/Acknowledgements.tex
\section*{Acknowledgements}
This work was supported by the National Natural Science Foundation of China (NSFC) under Grants 62472400 and 62271465, the National Key R\&D Program of China under Grant 2025YFC3408300, and the Suzhou Basic Research Program under Grant SYG202338.

%% file: Section/Appendices.tex
\appendix

\tcbset{
    promptstyle/.style={
        enhanced,
        breakable,
        colframe=gray!40!black,
        colback=gray!5,
        coltitle=white,
        fonttitle=\bfseries\small,
        attach boxed title to top left={yshift=-2mm, xshift=2mm},
        boxed title style={boxrule=0pt, colframe=gray!40!black, colback=gray!40!black, sharp corners},
        sharp corners,
        boxrule=0.8pt,
        top=8pt, bottom=8pt, left=8pt, right=8pt,
        before skip=10pt, after skip=10pt,
        fontupper=\small,
        drop shadow
    }
}

\section{Appendix}
\label{sec:appendix}

\subsection{Experiment Implementation Details}
\label{Experiment Detail}

\paragraph{Hardware \& Environment.}
All experiments were conducted on a server running Ubuntu 20.04.6 LTS, equipped with an Intel(R) Xeon(R) Platinum 8358 CPU @ 2.60GHz, 400GB of RAM, and four NVIDIA A100-80G GPUs. We implement our framework using PyTorch. To enable efficient similarity search, we utilize the FAISS library for vector indexing and retrieval.

\paragraph{Model Configuration.}
Table~\ref{tab:modelCard} details the architectures of the models employed in this study. For inference, all Large Language Models (LLMs) are served via the vLLM library to ensure high throughput.

\input{Appendix/model_card}

\paragraph{Baseline Settings.}
Detailed configurations for different experimental settings are as follows:

\begin{itemize}[leftmargin=*]
    \item \textbf{Zero-Shot:} We directly input the query and the image into the model to assess its inherent reasoning capabilities without external augmentation.
    
    \item \textbf{Chain of Thought (CoT):} We append the prompt ``Let's think step by step'' to the input to induce stepwise reasoning from the MLLM.

    \item \textbf{Fine-Tuning:} We employ the LLaMA-Factory framework to incorporate domain knowledge from the commonsense graph using the LoRA (Low-Rank Adaptation) method. The configuration includes a rank of 8, a scaling factor ($\alpha$) of 16, and a dropout rate of 0 applied to all linear layers. The training employs the AdamW optimizer with a learning rate of $5 \times 10^{-5}$ (cosine decay), gradient clipping at a maximum norm of 1.0, and BF16 mixed precision. To handle long contexts, we set the cutoff length to 2048, utilizing a per-device batch size of 2 with 8 steps of gradient accumulation.
    
    \item \textbf{Naive RAG:} We segment the reference text into chunks of 1024 tokens and utilize the \texttt{all-MiniLM-L6-v2} model for embedding generation. During retrieval, the top-3 chunks ($k=3$) are selected for context augmentation.
\end{itemize}

\subsection{Benchmark Details}
\label{Benchmark Detail}

\paragraph{FVQA2.0+}
This dataset is built upon FVQA 2.0, an extension of the original FVQA dataset designed for the Fact-based Visual Question Answering task. FVQA 2.0 tests a system's ability to answer visually related questions by leveraging external knowledge graphs, effectively addressing limitations such as small scale, imbalanced answer distribution, and overfitting. It comprises 2,820 QA pairs. On this basis, we propose \textbf{FVQA 2.0+}. We utilized an LLM-driven web search agent to perform visual modality augmentation on the textual knowledge graph of FVQA 2.0, constructing a Multimodal Knowledge Graph (MMKG) containing 1,152 nodes, 1,767 edges, and 3,342 images.

\paragraph{MVQA}
This benchmark is constructed based on MovieBench, a hierarchical dataset including high-level plot summaries and shot-level visual descriptions of classic movies. We selected 50 movies to construct an MMKG comprising 1,271 nodes, 1,468 edges, and 4,611 images. Furthermore, we employed an automated question generation pipeline to create 1,433 QA pairs centered on movie plots.
\begin{figure}[htbp]
    \centering
    \includegraphics[width=0.8\linewidth]{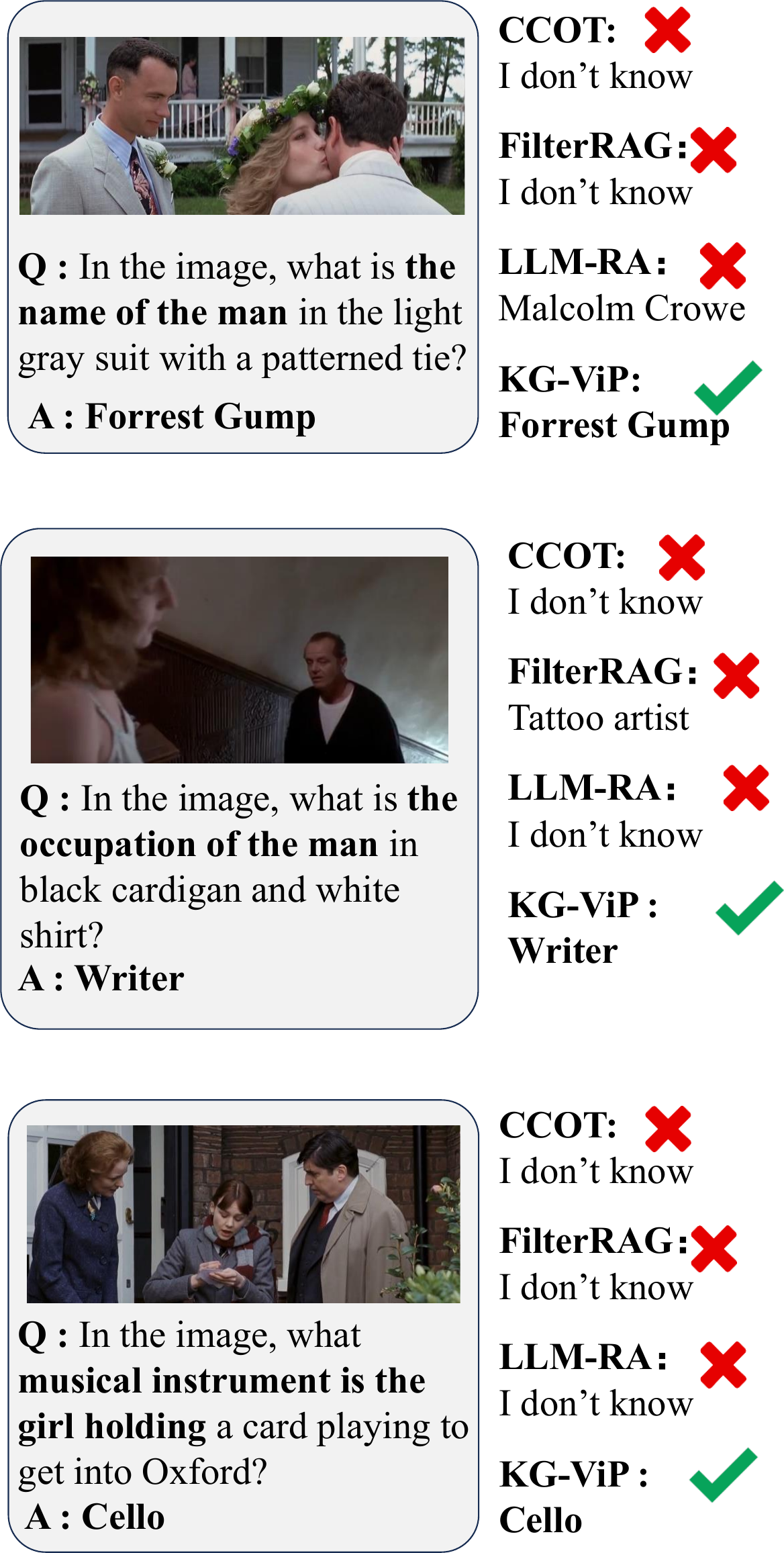}
    \caption{Qualitative comparison of VQA results. The visualization demonstrates the reasoning process and answer generation across different methods.}
    \label{fig:VQA_vis}
\end{figure}
\subsection{Case Study}
\paragraph{QA Generation Pipeline}
To alleviate the scarcity of manually annotated VQA pairs, we employ MLLMs to synthesize QA samples via prompt engineering. Specifically, the generation process is conditioned primarily on scene images and their textual descriptions to identify visible entities, interactions, and event cues. Concurrently, we incorporate an accessible commonsense graph as an external source of background knowledge to constrain the scope of knowledge dependency and enhance factual diversity (e.g., character identities, relationships, and plot-related common sense). In implementation, we retrieve local knowledge fragments associated with candidate entities from the graph and provide them as auxiliary context to the MLLM. This encourages the generation of questions that require reasoning over structured background knowledge beyond mere visually discernible content. Detailed prompt templates are provided in Appendix~\ref{Appendix Prompt}.

Figure~\ref{fig:VQA_vis} illustrates a qualitative comparison between our method and baseline approaches, including LLM-RA and FilterRAG. It can be observed that the latter are prone to generating ``hallucinated'' responses that appear plausible but are factually incorrect or ambiguous. In contrast, our proposed method consistently handles complex queries with high precision, particularly in scenarios requiring deep-level reasoning and multi-step knowledge integration.

\subsection{Prompt Templates}
\label{Appendix Prompt}

Here we present the key prompts used in our framework.

\begin{tcolorbox}[promptstyle, title={Scene Graph Generation}]
You are a visual grounding + relation extraction system.
Give you an Image. Your task is to extract key objects and relationships from this Image. Image size: width=\{w\}px, height=\{h\}px.
Return ONLY one valid JSON object.

\textbf{Rules:}
\begin{itemize}[leftmargin=*, nosep]
  \item Use \texttt{bbox\_px: [x1,y1,x2,y2]} in pixels.
  \item \texttt{0 <= x1 < x2 <= \{w\}}, \texttt{0 <= y1 < y2 <= \{h\}}.
  \item \texttt{confidence} in \texttt{[0,1]}. Omit uncertain ones.
\end{itemize}

\textbf{Schema:}
\begin{verbatim}
{
  "entities": [
    {
      "entity_id": "e1",
      "name": "person",
      "category": "person|object|animal|other",
      "bbox_px": [10, 20, 100, 200],
      "confidence": 0.90
    }
  ],
  "relations": [
    {
      "subject_id": "e1",
      "predicate": "next_to",
      "object_id": "e2",
      "confidence": 0.70,
      "evidence": "Two persons are adjacent."
    }
  ]
}
\end{verbatim}
\end{tcolorbox}

\begin{tcolorbox}[promptstyle, title={Query-guided Pruning}]
You are selecting which image-graph entities/relations should be kept for further commonsense graph retrieval. You can use the information in a text-guided commonsense subgraph.

\textbf{Text-guided commonsense subgraph:} \texttt{\{\}}
\textbf{Scene graph entities:} \texttt{\{\}}
\textbf{Scene graph relations:} \texttt{\{\}}

\textbf{Task:} Decide which scene graph entities and scene graph relations are relevant to the text-guided commonsense subgraph. Return ONLY valid JSON with schema:

\textbf{Schema:}
\begin{verbatim}
{
  "keep_entities": [
    "unique_name1", ...
  ],
  "keep_relations": [
    {
      "s": "unique_name1",
      "p": "predicate",
      "o": "unique_name2"
    }
  ]
}
\end{verbatim}
\end{tcolorbox}

\begin{tcolorbox}[promptstyle, title={LLM-as-a-Judge}]
You are an answer relevance evaluator. Given a reference answer and a model answer, your task is to output a single float score between 0 and 1 based on the following criteria:

1.0: Fully correct and complete semantic match.

0.8$\sim$0.9: Mostly correct, minor omissions or extras.

0.6$\sim$0.7: Correct core meaning but brief or incomplete.

0.3$\sim$0.5: Partially correct, captures only some aspects.

0.0$\sim$0.2: Incorrect or irrelevant.

Reference Answer: {reference}

Model Answer: {prediction}

Now output a single float score between 0 and 1, noting else!

\end{tcolorbox}

\begin{tcolorbox}[promptstyle, title={Graph Refinement}]
You are a knowledge graph analysis expert. Please analyze whether the current subgraph adequately answers the query question and provide optimization suggestions.

\textbf{Query:} ``\texttt{\{\}}''\\
\textbf{Query Image:} \texttt{\{\}}

\textbf{Current Subgraph Information:}
\begin{itemize}[leftmargin=*, nosep]
  \item Total Nodes: \texttt{\{\}}
  \item Total Edges: \texttt{\{\}}
\end{itemize}

Please analyze and provide optimization suggestions, focusing on:
\begin{itemize}[leftmargin=*, nosep]
  \item Does the subgraph adequately cover the key aspects of the query?
  \item Which nodes are core nodes that need to be further searched, and which may be noise nodes that need to be deleted?
\end{itemize}

Please respond strictly in the following JSON format, without any additional content:

\textbf{Schema:}
\begin{verbatim}
{
  "analysis": "Detailed analysis...",
  "recommendation": "Expand|Prune|Terminate",
  "nodes_to_expand": [],
  "nodes_to_prune": [],
  "reason": "Decision rationale",
  "confidence": 0.8
}
\end{verbatim}
\end{tcolorbox}

\begin{tcolorbox}[promptstyle, title={VQA Data Generator}]
You are a Visual Question Answering (VQA) data generator.

\textbf{Inputs:}
\begin{itemize}[leftmargin=*, nosep]
  \item (1) ONE scene image (the question image)
  \item (2) Scene description (characters, plot)
  \item (3) Reference facts (entities, attributes)
\end{itemize}

\textbf{Task:}
Generate one question grounded in the scene image and the scene description.
The question should not be fully answerable from the Image alone, and is expected to benefit from the provided reference facts (retrieved from a commonsense graph).

\textbf{Instructions:}
Look at the scene image first and choose a salient character (prefer one that is clearly visible).

\textbf{Guidelines (Person-centric questions):}
Try to involve at least one aspect that is typically not directly visible in the Image, such as name/identity, actor/portrays, occupation/role, condition/suffers from, ownership, or other personal attributes mentioned in the reference facts.

\textbf{Schema:}
\begin{verbatim}
{
  "qa_pairs": [
    {
      "Question": "...",
      "Answer": "..."
    }
  ]
}
\end{verbatim}

\vspace{2pt}
\textbf{Scene Image:} \texttt{\{\}}\\
\textbf{Scene description:} \texttt{\{\}}\\
\textbf{Reference facts:} \texttt{\{\}}
\end{tcolorbox}

%% file: Appendix/model_card.tex
\begin{table}[htbp]
\caption{Main Models Used in This Paper}
\label{tab:modelCard}
\centering
    \small
    \begin{tabular}{l|l} 
    \toprule
    Model Name & Base Architecture \\
    \midrule
    all-MiniLM-L6-v2           & BERT (Transformer Encoder) \\
    Cambrian-1-8b                & Llama-3 \\
    clip-vit-base-patch32      & ViT-B/32 \\
    blip-itm-base-coco         & ViT-B + BERT \\
    Qwen2.5-7B-Instruct        & Transformer Decoder \\
    Qwen2.5-VL-3B-Instruct             & Qwen2.5 + ViT \\
    Qwen2.5-VL-7B-Instruct     & Qwen2.5 + ViT \\
    Qwen2.5-VL-32B-Instruct             & Qwen2.5 + ViT \\
    GLM-4.1V-9B-Thinking              & GLM-4 + ViT \\
    GLM-4.5V-106B              & GLM-4.5 (MoE) + ViT \\
    \bottomrule
    \end{tabular}
\end{table}